\documentclass[draftclsnofoot,onecolumn]{IEEEtran}

\ifCLASSINFOpdf

\else

\fi
\usepackage{graphicx}
\usepackage{amsmath,amssymb} 
\usepackage{color}
\usepackage[noend]{algpseudocode}
\usepackage{mathrsfs}
\usepackage{bm}
\usepackage[noend]{algpseudocode}
\usepackage[ruled]{algorithm2e}
\usepackage{subfigure}
\usepackage{epstopdf}
\newcommand{\ie}{\emph{i.e.}}
\renewcommand{\footnoterule}{%
	\kern -2pt
	\hrule width 0.3\textwidth height 0.2pt
	\kern 2pt 
}
\begin{document}

\title{Deep Cross-modality Adaptation via Semantics Preserving Adversarial Learning for Sketch-based 3D Shape Retrieval}

\author{Jiaxin~Chen and~Yi~Fang
}

\markboth{}%
{Shell \MakeLowercase{\textit{et al.}}: Bare Demo of IEEEtran.cls for IEEE Journals}

\maketitle

\begin{abstract}
Due to the large cross-modality discrepancy between 2D sketches and 3D shapes, retrieving 3D shapes by sketches is a significantly challenging task. To address this problem, we propose a novel framework to learn a discriminative deep cross-modality adaptation model in this paper. Specifically, we first separately adopt two metric networks, following two deep convolutional neural networks (CNNs), to learn modality-specific discriminative features based on an importance-aware metric learning method. Subsequently, we explicitly introduce a cross-modality transformation network to compensate for the divergence between two modalities, which can transfer features of 2D sketches to the feature space of 3D shapes. We develop an adversarial learning based method to train the transformation model, by simultaneously enhancing the holistic correlations between data distributions of two modalities, and mitigating the local semantic divergences through minimizing a cross-modality mean discrepancy term. Experimental results on the SHREC 2013 and SHREC 2014 datasets clearly show the superior retrieval performance of our proposed model, compared to the state-of-the-art approaches.
\end{abstract}


\section{Introduction}


In the last few years, there has been an explosive growth of 3D shape data, due to increasing demands from real industrial applications, such as virtual reality, LiDAR based autonomous vehicles. 3D shape related techniques have emerged as extremely hot research topics recently. Retrieving a certain category of 3D shapes from a given database is one of the fundamental problems for 3D shape based applications. A lot of efforts have been devoted to 3D shape retrieval by 3D models \cite{wang2017cnn,xie2017deepshape}, which are intuitively straightforward, but difficult to acquire. Alternatively, freehand sketch is a more convenient way for human to
interact with data collection and processing systems, especially with the sharply increased use of touch-pad devices such as smart phones and tablet computers. As a consequence, sketch-based 3D shape retrieval, \ie, searching 3D shapes queried by sketches, has attracted more and more attentions \cite{eitz2012sketch,wang2015sketch,li2015comparison,CVPR2017Xie}.

\begin{figure}[!t]
\centering
\includegraphics[height=8.1cm]{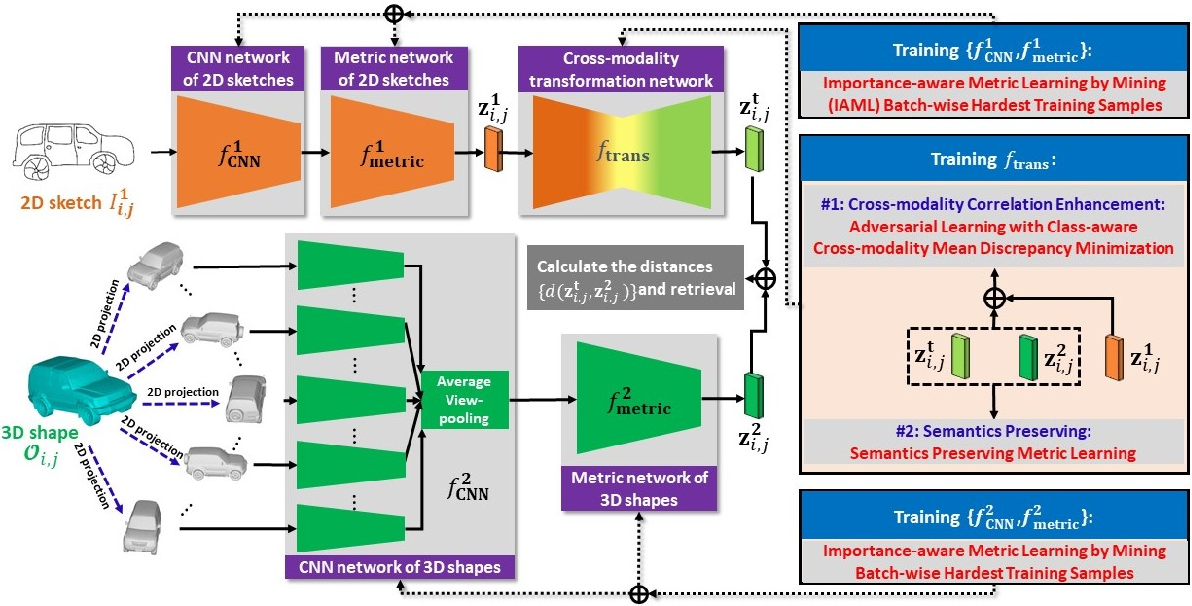}
\caption{Framework of our proposed method. Our model consists of the CNN network $f^{\textbf{1}}_{\textrm{\textbf{CNN}}}$ and metric network $f^{\textbf{1}}_{\textrm{\textbf{metric}}}$ of 2D sketches, the CNN network $f^{\textbf{2}}_{\textrm{\textbf{CNN}}}$ and metric network $f^{\textbf{2}}_{\textrm{\textbf{metric}}}$ of rendered images of 3D shapes, together with the cross-modality transformation network $f_{\textrm{\textbf{trans}}}$. The CNN and metric networks for each single modality (\ie, 2D sketches or 3D shapes) is trained by importance-aware metric learning through mining the hardest training samples. The cross transformation network $f_{\textrm{\textbf{trans}}}$ is trained by enforcing features of sketches to be semantics preserving after transform. Simultaneously, an adversarial learning with cross-modality mean discrepancy minimization is employed to enhance both the local and holistic correlations between data distributions of transformed features of sketches and features of 3D shapes. }
\label{fig:framework}
\vspace{-0.1in}
\end{figure}

Despite of its succinctness and convenience to acquire, freehand sketches remain two disadvantages in the application of 3D shape retrieval, making the sketch-based 3D shape retrieval an extremely challenging task. Firstly, sketches are usually drawn subjectively in uncontrolled environments, resulting in \emph{severe intra-class variations} as shown in Fig. \ref{fig:dataset}. Secondly, sketches and 3D shapes have heterogenous data structures, which leads to \emph{large cross-modality divergences}.

A variety of models have been proposed to address the aforementioned two issues, which can be roughly divided into two categories, \ie, representation based methods and matching based methods. The first category aims to extract robust features for both sketches and 3D shapes \cite{li2013shrec,li2015comparison,eitz2012sketch,furuya2013ranking,wang2015sketch,zhu2016learning,zhu2016heat,CVPR2017Xie}. However, due to the heterogeneity of sketches and 3D shapes, it is quite difficult to achieve modality-invariant discriminative representations. On the other hand, matching based methods focus on developing effective models for calculating similarities or distances between sketches and 3D shapes, among which deep metric learning based models \cite{wang2015sketch,dai2017deep,CVPR2017Xie} have achieved the state-of-the-art performance. Nevertheless, these methods fail to explore the varying importance of different training samples. Besides, they can merely enhance local cross-modality correlations, by selecting data pairs or triplets across modalities, while not taking into account the holistic data distributions. As a consequence, the learned deep metrics might be less discriminative, and lack of generalization for unseen test data.

To overcome the drawbacks of existing works, we propose a novel model, namely Deep Cross-modality Adaptation (DCA), for sketch-based 3D shape retrieval. Fig. \ref{fig:framework} shows the framework of our proposed model. We first construct two separate deep convolutional neural networks (CNNs) and metric networks, one for sketches and the other for 3D shapes, to learn discriminative modality-specific features for each modality via importance-aware metric learning (IAML). Through mining the hardest samples in each mini-batch for training, IAML could explore the importance of training data, and therefore learn discriminative representations more efficiently. Furthermore, in order to reduce the large cross-modality divergence between learned features of sketches and 3D shapes, we explicitly introduce a cross-modality transformation network, to transfer features of sketches into the feature space of 3D shapes. An adversarial learning method with class-aware cross-modality mean discrepancy minimization (CMDM-AL) is developed to train the transformation network, which acts as a generator. Since CMDM-AL is able to enhance correlations between distributions of transferred data of sketches and data of 3D shapes, our model can compensate for the cross-modality discrepancy in a holistic way. IAML is also applied to the transformed data, in order to further preserve semantic structures of sketch data after adaptation. The main contributions of this paper are three-fold:

1) We propose a novel deep cross-modality adaptation model via semantics preserving adversarial learning. To our best knowledge, this work is the first one that incorporates adversarial learning into sketch-based 3D shape retrieval.


2) We develop a new adversarial learning based method for training the deep cross-modality adaptation network, which simultaneously reduces the holistic cross-modality discrepancy of data distributions, and enhances semantic correlations of local data batches across modalities.


3) We significantly boost the performance of existing state-of-the-art sketch-based 3D shape retrieval methods on two large benchmark datasets.

\section{Related Work}

In the literature, most of existing works on sketch-based 3D shape retrieval mainly concentrate on building modality-invariant representations for sketches and 3D shapes, and developing discriminative matching models. Various hand-craft features are employed, such as Zernike moments, coutour-based Fourier descriptor, eccentricity feature and circularity feature \cite{li2014shrec}, the chordal axis transform based shape descriptor \cite{yasseen2017view}, HoG-SIFT features \cite{yoon2017sketch}, the local improved Pyramid of Histograms of Orientation Gradients (iPHOG) \cite{li2017sketch}, the sparse coding spatial pyramid matching feature (ScSPM), local depth scale-invariant feature transform (LD-SIFT) \cite{zhu2016learning}. Besides, many learning-based features are developed, including bag-of-features (BoF) with Gabor local line based features (GALIF) \cite{li2014comparison}, dense SIFT with BOF \cite{furuya2013ranking}. Meanwhile, tremendous matching approaches have also been developed, such as manifold ranking \cite{furuya2013ranking}, dynamic time warping \cite{yasseen2017view}, sparse coding based matching \cite{yoon2017sketch} and adaptive view clustering \cite{li2013shrec,li2017sketch}.

Recently, various deep models have been developed for both feature extraction and matching, which are closely related to our proposed method. In \cite{wang2015sketch}, two Siamese CNNs were employed to learn discriminative features of sketches and 3D shapes by minimizing within-modality and cross-modality losses. In \cite{zhu2016learning}, the pyramid cross-domain neural networks were utilized to compensate for cross-domain divergences. In \cite{dai2017deep} and \cite{xie2017deepshape}, Siamese metric networks are employed to minimize both within-modality and cross-modality intra-class distances whilst maximizing inter-class distances. In \cite{xie2017deepshape}, the Wasserstein barycenters were additionally employed to aggregate multi-view deep features of rendered images from 3D models. However, these methods only reduced the local cross-modality divergence, and haven't considered  removing data distribution shift across modalities. In contrast, our proposed model employs an adversarial learning based method to mitigate the discrepancy between distributions of two modalities in a holistic way, whilst addressing the local divergence issues by introducing a class-aware mean discrepancy term. Moreover, we apply IAML to mine importance of training samples, which has also been ignored by current works.

Another branch of works related to our work is the supervised discriminative adversarial learning for domain adaptation. In \cite{ganin2016domain,tzeng2017adversarial,motiian2017few}, a variety of adversarial discriminative models were developed for domain adaptation. The basic idea of these methods is to remove the domain shift between the source and target domains, by employing a domain discriminator and an adversarial loss. However, these works concentrate on scenarios where few labeled data are available in the target domain (despite abundant labeled data in the source domain), and are unable to jointly explore local discriminative semantic structures for both domains, making them unsuitable for our task. In \cite{zhang2017aspect}, the authors also explicitly adopted a transformation network to transfer data from source domain to the target domain, where the cross-domain divergence is mitigated by an adversarial loss. However, they used hand crafted features, while our model employs deep CNNs to learn discriminative modality-specific features, and integrates them with the transformation network as a whole. Moreover, we introduce a class-aware cross-modality mean discrepancy term to the original adversarial loss. This term can enhance semantic correlations of data distributions across modalities as well as remove domain shift, which is largely neglected by existing works.


\section{Deep Cross-modality Adaptation}
\label{sec:proposed}

As illustrated in Fig. \ref{fig:framework}, our proposed framework mainly consists of five components, including the CNN networks for 2D sketches (denoted by $f^{\textbf{1}}_{\textbf{\textrm{CNN}}}$) and for 3D shapes (denoted by $f^{\textbf{2}}_{\textrm{\textbf{CNN}}}$), fully connected metric networks for 2D sketches (denoted by $f^{\textbf{1}}_{\textrm{\textbf{metric}}}$) and for 3D shapes (denoted by $f^{\textbf{2}}_{\textrm{\textbf{metric}}}$), together with the cross-modality transformation network $f_{\textrm{\textbf{trans}}}$, of which the parameters are $\bm{\theta}^{\textbf{1}}_{\textrm{\textbf{CNN}}}$, $\bm{\theta}^{\textbf{2}}_{\textrm{\textbf{CNN}}}$, $\bm{\theta}^{\textbf{1}}_{\textrm{\textbf{metric}}}$, $\bm{\theta}^{\textbf{2}}_{\textrm{\textbf{metric}}}$ and $\bm{\theta}_{\textrm{\textbf{trans}}}$, respectively.

Similar to most existing deep learning methods, we train our model by mini-batches. In order to depict our own method more conveniently, we build image batches from the whole training data in a slightly different way from random sampling. Specifically, for 2D sketches, we first select $C$ classes randomly, and then collect $K$ images for each class. The selected images finally comprise a mini-batch $\bm{\mathcal{I}}^{\textbf{1}}=\left\{I_{1,1}^{\textbf{1}}, \cdots, I_{1,K}^{\textbf{1}}, \cdots, I_{C,1}^{\textbf{1}}, \cdots, I_{C,K}^{\textbf{1}}\right\}$ of size $C\times K$, of which the corresponding class labels are denoted by $\textbf{Y}^{\textbf{1}}=\left\{y_{1}, \cdots, y_{1}, \cdots, y_{C}, \cdots, y_{C}\right\}$. Following the same way, a batch of 3D shapes $\bm{\mathcal{O}}=\left\{O_{1,1}, \cdots, O_{1,K}, \cdots, O_{C,1}, \cdots, O_{C,K}\right\}$ is constructed, together with labels $\textbf{Y}^{\textbf{2}}=\left\{y_{1}, \cdots, y_{1}, \cdots, y_{C}, \cdots, y_{C}\right\}$. To characterize a 3D shape, we utilize the widely used multi-view representation as in \cite{su2015multi,dai2017deep,CVPR2017Xie}, \ie, projecting a 3D shape to $N_{v}$ grayscale images from $N_{v}$ rendered views that are evenly divided around the 3D shape. Thereafter, we can represent $\bm{\mathcal{O}}$ as a batch of images $\bm{\mathcal{I}}^{\textbf{2}}=\left\{\textbf{I}_{1,1}^{\textbf{2}}, \cdots, \textbf{I}_{1,K}^{\textbf{2}}, \cdots, \textbf{I}_{C,1}^{\textbf{2}}, \cdots, \textbf{I}_{C,K}^{\textbf{2}}\right\}$, where $\textbf{I}_{i,j}^{\textbf{2}}=\left\{I_{i,j,v}^{\textbf{2}}\right\}_{v=1}^{N_{v}}$ consists of $N_{v}$ ($N_{v}$=12 is used in our paper) 2D rendered images of the 3D shape $O_{i,j}$.

As demonstrated in Fig. \ref{fig:framework}, we train the CNN and metric networks for sketches, \ie, $f^{\textbf{1}}_{\textbf{\textrm{CNN}}}$ and $f^{\textbf{1}}_{\textbf{\textrm{metric}}}$, jointly by adopting an importance-aware metric learning method, which could explore hardest training samples within a mini-batch. The CNN and metric networks for 3D shapes, \ie, $f^{\textbf{2}}_{\textbf{\textrm{CNN}}}$ and $f^{\textbf{2}}_{\textbf{\textrm{metric}}}$, are also trained in the same way. The cross-modality transformation network $\bm{\theta}_{\textrm{\textbf{trans}}}$ is learnt by preserving semantic structures of transformed features, and employing an adversarial learning based training strategy with class-aware cross-modality mean discrepancy minimization.

In the rest of this paper, we will elaborate the trailing details about the proposed method, including the importance-aware metric learning, the semantic adversarial learning, and the optimization algorithm. Without loss of generality, all loss functions are formulated based on image batches $\bm{\mathcal{I}}^{\textbf{1}}$ and $\bm{\mathcal{I}}^{\textbf{2}}$ throughout this paper, which can be easily extended to the whole training data.


\subsection{Importance-Aware Feature Learning}\label{subsec:SMML}


Given a mini-batch $\bm{\mathcal{I}}^{m}$, after successively passing $\bm{\mathcal{I}}^{m}$ through the CNN network $f^{m}_{\textrm{\textbf{CNN}}}$ and the metric network $f_{\textrm{\textbf{metric}}}^{m}$, we can obtain a set of feature vectors:
$$
\textbf{Z}^{m}=\left\{\textbf{z}_{1,1}^{m}, \cdots, \textbf{z}_{1,K}^{m}, \cdots, \textbf{z}_{C,1}^{m}, \cdots, \textbf{z}_{C,K}^{m}\right\},
$$
where $m \in \{\textbf{1},\textbf{2}\}$, and for $i=1,\cdots, C$, $j=1,\cdots,K$
$$
\textbf{z}_{i,j}^{\textbf{1}} = f_{\textrm{\textbf{metric}}}^{\textbf{1}}\left(f^{\textbf{1}}_{\textrm{\textbf{CNN}}}(I_{i,j}^{\textbf{1}})\right), \textbf{z}_{i,j}^{\textbf{2}} = f_{\textrm{\textbf{metric}}}^{\textbf{2}}\left(f^{\textbf{2}}_{\textrm{\textbf{CNN}}}(\textbf{I}_{i,j}^{\textbf{2}})\right).
$$

Ideally, in order to learn discriminative features for each modality (\ie, the 2D sketches or the 3D shapes), the inter-class distances within the batch $\textbf{Z}^{m}$ need to be larger than the intra-class distances. To achieve this, inspired by \cite{hermans2017defense}, we adopt the following loss function for importance-aware metric learning:
\begin{equation}\label{eq:loss_metric_learning}
\begin{aligned}
  & L^{m}_{IAML}(\left\{\bm{\theta}^{m}_{\textrm{\textbf{CNN}}},\bm{\theta}^{m}_{\textrm{\textbf{metric}}}\right\};\textbf{Z}^{m}) \\& = \sum_{i}^{C}\sum_{j=1}^{K} \max\left( 0, \eta-\left[\left\|\textbf{z}_{i,j}^{m}-\textbf{z}_{i^{*},n^{*}}^{m}\right\|_{2}-\left\|\textbf{z}_{i,j}^{m}-\textbf{z}_{i,p^{*}}^{m}\right\|_{2}\right] \right),
\end{aligned}
\end{equation}
where
\begin{equation}\label{eq:dist_farthest_inter_class}
 \textbf{z}_{i^{*},j^{*}}^{m} = \mathop{\rm{argmin}}\limits_{i^{'}\in \{1,\cdots,C\},y_{i^{'}}\neq y_{i},n\in \{1,\cdots,K\} }\left\|\textbf{z}_{i,j}^{m}-\textbf{z}_{i^{'},n}^{m}\right\|_{2},
\end{equation}
\begin{equation}\label{eq:dist_nearest_inter_class}
 \textbf{z}_{i,p^{*}}^{m} = \mathop{\rm{argmax}}\limits_{p\in \{1,\cdots,K\}, p\neq j }\left\|\textbf{z}_{i,j}^{m}-\textbf{z}_{i,p}^{m}\right\|_{2},
\end{equation}
and $\eta>0$ is a constant.


As can be seen from Eq. \eqref{eq:dist_farthest_inter_class}, for a certain anchor point $\textbf{z}_{i,j}^{m}$, $\textbf{z}_{i^{*},n^{*}}^{m}\in \textbf{Z}^{m}$ is the sample that has the minimal Euclidean distance to $\textbf{z}_{i,j}^{m}$ among those samples from different classes. And from Eq. \eqref{eq:dist_nearest_inter_class}, we can see that $\textbf{z}_{i,p^{*}}^{m}$ is the sample that has the maximal Euclidean distance to $\textbf{z}_{i,j}^{m}$, among samples belonging to the same class as $\textbf{z}_{i,j}^{m}$. In other words, $\left\|\textbf{z}_{i,j}^{m}-\textbf{z}_{i^{*},n^{*}}^{m}\right\|_{2}$ and $\left\|\textbf{z}_{i,j}^{m}-\textbf{z}_{i,p^{*}}^{m}\right\|_{2}$ indicate the largest inter-class Euclidean distance and the minimal intra-class Euclidean distance with respect to $\textbf{z}_{i,j}^{m}$ within the batch $\textbf{Z}^{m}$, respectively. Therefore, $\textbf{z}_{i,p^{*}}^{m}$ and $\textbf{z}_{i^{*},n^{*}}^{m}$ are the batch-wise ``\emph{hardest positive}" and the ``\emph{hardest negative}" samples w.r.t. $\textbf{z}_{i,j}^{m}$, and should be given higher importance during training. Existing deep metric learning based models \cite{dai2017deep,xie2017deepshape} equally treat all training samples. In contrast, our IAML firstly explores the hardest positive and negative training samples within a mini-batch, and enforces them to be consistent with semantics, making it more efficient to learn discriminative features.

By minimizing $L^{m}_{IAML}$ in Eq. \eqref{eq:loss_metric_learning}, $\left\|\textbf{z}_{i,j}^{m}-\textbf{z}_{i^{*},n^{*}}^{m}\right\|_{2}-\left\|\textbf{z}_{i,j}^{m}-\textbf{z}_{i,p^{*}}^{m}\right\|_{2}$
are forced to be greater than $\eta$, \ie, $\left\|\textbf{z}_{i,j}^{m}-\textbf{z}_{i^{*},n^{*}}^{m}\right\|_{2}-\left\|\textbf{z}_{i,j}^{m}-\textbf{z}_{i,p^{*}}^{m}\right\|_{2}>\eta$. That is to say, by minimizing $L^{m}_{IAML}$, the minimal inter-class distance is compelled to be larger than the maximal intra-class distance in the feature space, whilst keeping a certain margin $\eta$. Consequently, we can learn CNN and metric networks to extract discriminative features for each modality (\ie, 2D sketches or 3D shapes).

\subsection{Cross-modality Transformation based on Adversarial Learning}

By applying the importance-aware metric learning via minimizing the losses $L^{\textbf{1}}_{SMML}$ and
$L^{\textbf{2}}_{SMML}$, we can learn discriminative features for sketches and shapes, \ie, $\{\textbf{z}_{i,j}^{\textbf{1}}\}$ and $\{\textbf{z}_{i,j}^{\textbf{2}}\}$, respectively. However, due to the large discrepancy between data distributions of different modalities, directly using $\{\textbf{z}_{i,j}^{\textbf{1}}\}$ and $\{\textbf{z}_{i,j}^{\textbf{2}}\}$ for cross-modality retrieval will result in extremely poor performance.

To address this problem, we propose a cross-modality transformation network $f_{\textrm{\textbf{trans}}}$, in order to adapt the learnt features of 2D sketches to the feature space of 3D shapes with cross-modality discrepancies removal.

Suppose $\textbf{Z}^{\textbf{t}}=\{\textbf{z}_{i,j}^{\textbf{t}}\}$ is the transformed features of sketches $\textbf{Z}^{\textbf{1}}=\{\textbf{z}_{i,j}^{\textbf{1}}\}$ with class labels 
$$\textbf{Y}^{\textbf{t}}=\left\{y_{1}, \cdots, y_{1}, \cdots, y_{C}, \cdots, y_{C}\right\},$$ 
where $\textbf{z}_{i,j}^{\textbf{t}}=f_{\textrm{\textbf{trans}}}(\textbf{z}_{i,j}^{\textbf{1}}|\bm{\theta}_{\textrm{\textbf{trans}}})$ for $\forall i\in \{1,\cdots,C\}$, and $j\in \{1,\cdots,K\}$. Ideally, the transformed features $\{\textbf{z}_{i,j}^{\textbf{t}}\}$ are expected to have the following properties, in order to guarantee good performance for the cross-modality retrieval task:

1) $\{\textbf{z}_{i,j}^{\textbf{t}}\}$ should be \emph{semantics preserving}, \ie, maintaining small intra-class distances and large inter-class distances.

2) $\{\textbf{z}_{i,j}^{\textbf{t}}\}$ should have \emph{correlated data distribution} with $\{\textbf{z}_{i,j}^{\textbf{2}}\}$, \ie, the learnt features of 3D shapes.

\begin{figure}[!t]
\centering
\includegraphics[height=9.1cm]{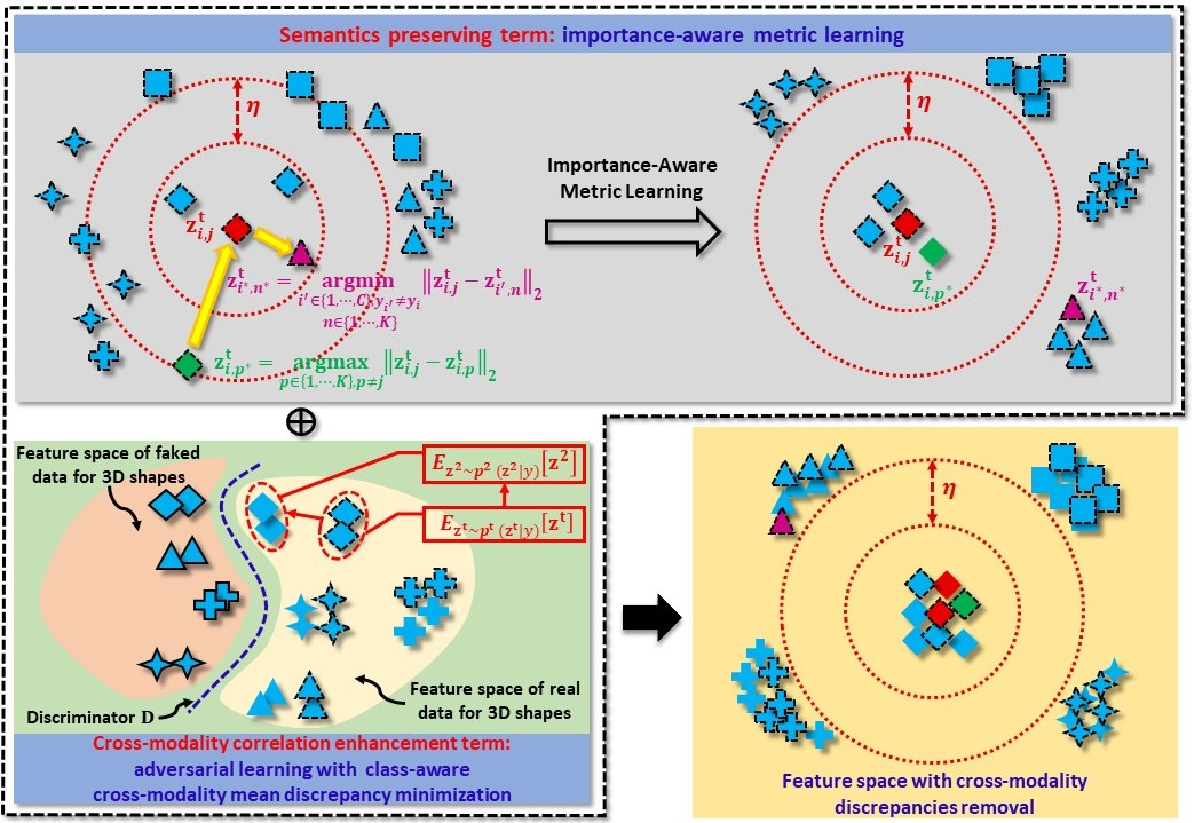}
\caption{Illustration on training the cross-modality transformation network. Importance-aware Metric Learning is applied to the transformed features of sketches to preserve semantic structures. An adversarial learning based method is developed to reduce the divergence between distributions of two modalities (\ie, sketches and 3D shapes). A class-aware mean discrepancy term is simultaneously minimized to further strengthened correlations between local batch-wise features across modalities. Here, shapes with solid (dashed) bounding boxes indicate faked data of 3D shapes (transformed sketch data). Shapes without bounding boxes indicate real data of 3D shapes.)}\label{fig:illustrate_cross_modality_metric_learning}
\vspace{-0.1in}
\end{figure}

The first property aims to compel the transformed features to preserve semantics, whilst the second attempts to remove the cross-modality discrepancy through strengthening correlations between data distributions of two modalities.

As shown in Fig. \ref{fig:illustrate_cross_modality_metric_learning}, we introduce a semantics preserving term by repeatedly utilizing the importance-aware metric learning to accomplish 1). And in order to achieve 2), we employ a cross-modality correlation enhancement term based on adversarial learning with class-aware cross-modality mean discrepancy minimization. We will provide details about the aforementioned two terms in the rest of this section.

\noindent \textbf{Semantics Preserving Term}
In order to preserve semantic structures, \ie, keeping small (large) intra-class (inter-class) distances, we apply the loss of Importance-aware Metric Learning previously introduced to transformed data:
\begin{equation}\label{eq:loss_adapt_SP}
L_{SeP}\left(\bm{\theta}_{\textrm{\textbf{trans}}}\right)= \sum_{i=1}^{C}\sum_{j=1}^{K} \max\left( 0, \eta-\left[\left\|\textbf{z}_{i,j}^{\textbf{t}}-\textbf{z}_{i^{*},n^{*}}^{\textbf{t}}\right\|_{2}-\left\|\textbf{z}_{i,j}^{\textbf{t}}-\textbf{z}_{i,p^{*}}^{\textbf{t}}\right\|_{2}\right] \right),
\end{equation}
where
\begin{equation}\label{eq:adapt_SP_dist_farthest_inter_class}
 \textbf{z}_{i^{*},n^{*}}^{\textbf{t}} = \mathop{\rm{argmin}}\limits_{i^{'}\in \{1,\cdots,C\},y_{i^{'}}\neq y_{i},n\in \{1,\cdots,K\} }\left\|\textbf{z}_{i,j}^{\textbf{t}}-\textbf{z}_{i^{'},n}^{\textbf{t}}\right\|_{2},
\end{equation}
\begin{equation}\label{eq:adapt_SP_dist_nearest_inter_class}
 \textbf{z}_{i,p^{*}}^{\textbf{t}} = \mathop{\rm{argmax}}\limits_{p\in \{1,\cdots,K\}, p\neq j }\left\|\textbf{z}_{i,j}^{\textbf{t}}-\textbf{z}_{i,p}^{\textbf{t}}\right\|_{2},
\end{equation}
and $\eta>0$ is a constant.

\noindent \textbf{Cross-modality Correlation Enhancement Term}
Generative adversarial networks (GANs) have recently emerged as an effective method to generate synthetic data \cite{goodfellow2014generative}. The basic idea is to train two competing networks, a generator $G$ and a discriminator $D$, based on game theory. The generator $G$ is trained to sample from the data distribution $p_{\textbf{x}}(\textbf{x})$ from the vector of noise $\textbf{v}$. The discriminator $D$ is trained to distinguish synthetic data generated by $G$ and real data sampled from $p_{\textbf{x}}(\textbf{x})$. The problem of training GANs is  formulated as follows:
\begin{equation}\label{eq:origGAN}
\min_{G}\max_{D} L_{GAN}:=E_{\textbf{x}\sim p_{\textbf{x}}(\textbf{x})}\left[\log(D({\textbf{x}}))\right]+E_{\textbf{v}\sim p_{\textbf{v}}(\textbf{v})}\left[\log(1-D(G(\textbf{v})))\right],
\end{equation}
where $p_{\textbf{v}}(\textbf{v})$ is a prior distribution over $\textbf{v}$. It has been pointed out in \cite{goodfellow2014generative} that the global equilibrium of the two-player game in Eq. \eqref{eq:origGAN} achieves if and only if $p_{\textbf{x}}(\textbf{x}) = p_{g}(\textbf{x})$, where $p_{g}(\textbf{x})$ is the distribution of generated data.

In our model, we treat the transformation network $f_{\textrm{\textbf{trans}}}$ as the generator $G$. Suppose $p^{\textbf{1}}(\textbf{z}^{\textbf{1}})$, $p^{\textbf{2}}(\textbf{z}^{\textbf{2}})$ and $p^{\textbf{t}}(\textbf{z}^{\textbf{t}})$ are distributions of learnt features of sketches, 3D shapes and transformed data (denoted by  $\textbf{z}^{\textbf{1}}$, $\textbf{z}^{\textbf{2}}$ and $\textbf{z}^{\textbf{t}}$), respectively. By solving the following problem
\begin{equation}\label{eq:adpatGAN}
\min_{f_{\textrm{\textbf{trans}}}}\max_{D} E_{\textbf{z}^{\textbf{2}}\sim p^{\textbf{2}}(\textbf{z}^{\textbf{2}})}\left[\log(D({\textbf{z}^{\textbf{2}}}))\right]+E_{\textbf{z}^{\textbf{1}}\sim p^{\textbf{1}}(\textbf{z}^{\textbf{1}})}\left[\log(1-D(f_{\textrm{\textbf{trans}}}(\textbf{z}^{\textbf{1}})))\right],
\end{equation}
we can expect that $p^{\textbf{t}}(\textbf{z}^{\textbf{t}})=p^{\textbf{t}}(f_{\textrm{\textbf{trans}}}(\textbf{z}^{\textbf{1}}))=p^{\textbf{2}}(\textbf{z}^{\textbf{2}})$, \ie, the transformed data $\textbf{z}^{\textbf{t}}$ has the same data distribution as $\textbf{z}^{\textbf{2}}$ of 3D shapes, if problem \eqref{eq:adpatGAN} reaches the global equilibrium. Consequently, the cross-modality discrepancy can be reduced.

Conventionally, problem \eqref{eq:adpatGAN} is solved by alternatively optimizing $f_{\textrm{\textbf{trans}}}$ and $D$ through minimizing the following two loss functions:
\begin{equation}\label{eq:loss_G}
L_{G}=E_{\textbf{z}^{\textbf{1}}\sim p^{\textbf{1}}(\textbf{z}^{\textbf{1}})}\left[\log(1-D(\textbf{z}^{\textbf{t}})))\right],
\end{equation}
\begin{equation}\label{eq:loss_D}
L_{D}=-E_{\textbf{z}^{\textbf{2}}\sim p^{\textbf{2}}(\textbf{z}^{\textbf{2}})}\left[\log(D({\textbf{z}^{\textbf{2}}}))\right]-E_{\textbf{z}^{\textbf{1}}\sim p^{\textbf{1}}(\textbf{z}^{\textbf{1}})}\left[\log(1-D(\textbf{z}^{\textbf{t}}))\right].
\end{equation}

So far, we have trained a transformation network $f_{\textrm{\textbf{trans}}}$ such that $p^{\textbf{t}}(\textbf{z}^{\textbf{t}})\approx p^{\textbf{2}}(\textbf{z}^{\textbf{2}})$ by minimizing $L_{G}$ and $L_{D}$. Albeit the divergence between the distributions for transformed features of sketches and for features of 3D models can be diminished by adversarial learning, the cross-modality semantic structures are not taken into account. To address this problem, we further introduce the following term, namely the class-aware cross-modality mean discrepancy
\begin{equation}\label{eq:loss_CMD}
L_{CMD}=\sum_{y}\left\|E_{\textbf{z}^{\textbf{t}}\sim p^{\textbf{t}}(\textbf{z}^{\textbf{t}}|y)}\left[\textbf{z}^{\textbf{t}}\right]-E_{\textbf{z}^{\textbf{2}}\sim p^{\textbf{2}}(\textbf{z}^{\textbf{2}}|y)}\left[\textbf{z}^{\textbf{2}}\right]\right\|_{2},
\end{equation}
to adversarial learning, where $y$ is the class label. By minimizing $L_{CMD}$, the mean feature vector of class $y$ from the sketch modality is compelled to be close to the mean feature vector of the same class from the 3D shape modality.

In practice, provided a mini-batch $\textbf{Z}^{q}=\{\textbf{z}_{i,j}^{q}\}_{i=1,j=1}^{C,K}$ ($q\in \{\textbf{2},\textbf{t}\}$), the term $E_{\textbf{z}^{q}\sim p^{q}(\textbf{z}^{q}|y)}\left[\textbf{z}^{q}\right]$ can be approximated by the batch-wise mean feature vector, \ie, $E_{\textbf{z}^{q}\sim p^{q}(\textbf{z}^{q}|y)}\left[\textbf{z}^{q}\right]\approx\frac{1}{K}\sum_{j=1,c_{i}=y}^{K}\textbf{z}_{i,j}^{q}$.

Through minimizing the loss $L_{AL}=L_{G}+L_{CMD}$, we can obtain the adversarial learning method with cross-modality mean discrepancy minimization (CMDM-AL), which could enhance the semantic correlations across modalities.

By combing the semantics preserving loss $L_{SeP}$ and the cross-modality correlation enhancing loss $L_{AL}$, we finally get the loss function for training $f_{\textrm{\textbf{trans}}}$:
\begin{equation}\label{eq:loss_adapt}
L_{T}(\bm{\theta}_{\textrm{\textbf{trans}}})=L_{SeP}+(L_{G}+L_{CMD}).
\end{equation}

\begin{algorithm}[!t]
\caption{Training the Deep Cross-modality Adaptation Model}\label{algo:CMT}
\LinesNumbered
\KwIn{2D sketches $\bm{\mathcal{I}}^{\textbf{1}}$, and rendered images of 3D shapes $\bm{\mathcal{I}}^{\textbf{2}}$.}
\KwOut{Trained parameters $\left\{\bm{\theta}_{\textrm{\textbf{CNN}}}^{\textbf{1}}, \bm{\theta}_{\textrm{\textbf{metric}}}^{\textbf{1}}\right\}$ for CNN and metric networks of sketches, $\left\{\bm{\theta}_{\textrm{\textbf{CNN}}}^{\textbf{2}},\bm{\theta}_{\textrm{\textbf{metric}}}^{\textbf{2}}\right\}$ for CNN and metric networks of 3D shapes, and $\bm{\theta}_{\textrm{\textbf{trans}}}$ for the cross-modality transformation network. }
\textbf{Initialization:} Pre-train $\{\bm{\theta}_{\textrm{\textbf{CNN}}}^{\textbf{1}}, \bm{\theta}_{\textrm{\textbf{metric}}}^{\textbf{1}}\}$ and $\{\bm{\theta}_{\textrm{\textbf{CNN}}}^{\textbf{2}}, \bm{\theta}_{\textrm{\textbf{metric}}}^{\textbf{2}}\}$, by minimizing $L_{IAML}^{\textbf{1}}$ and $L_{SMML}^{\textbf{2}}$ in Eq. \eqref{eq:loss_metric_learning}, respectively. Pre-train $\bm{\theta}_{\textrm{\textbf{trans}}}$ by minimizing $L_{T}$ and $L_{D}$. Set the maximal number of iteration steps$Itr_{\max}$, and the iterative step $k:=1$.\\
\While{$k\leq Itr_{\max}$}{
    Update $\left\{\bm{\theta}_{\textrm{\textbf{CNN}}}^{\textbf{1}}, \bm{\theta}_{\textrm{\textbf{metric}}}^{\textbf{1}}\right\}$ by reducing $L^{\textbf{1}}_{IAML}$ in Eq. \eqref{eq:loss_metric_learning}.\\
    Update $\left\{\bm{\theta}_{\textrm{\textbf{CNN}}}^{\textbf{2}}, \bm{\theta}_{\textrm{\textbf{metric}}}^{\textbf{2}}\right\}$ by reducing $L^{\textbf{2}}_{IAML}$ in Eq. \eqref{eq:loss_metric_learning}.\\
    Update the adversarial discriminator $D$ by optimizing $L_{D}$ in Eq. \eqref{eq:loss_D}.\\
    Update $\bm{\theta}_{\textrm{\textbf{trans}}}$ by reducing $L_{T}$ in Eq. \eqref{eq:loss_adapt}.\\
    $k:=k+1$.
}
\end{algorithm}
\vspace{-0.1in}

\subsection{Optimization}

In Eq. \eqref{eq:loss_metric_learning}, we defined the loss function $L^{\textbf{1}}_{IAML}$ for jointly training $\left\{f^{\textbf{1}}_{\textbf{\textrm{CNN}}}(\bm{\theta}_{\textrm{\textbf{CNN}}}^{\textbf{1}})\right.$, $\left.f^{\textbf{1}}_{\textbf{\textrm{metric}}}(\bm{\theta}_{\textrm{\textbf{metric}}}^{\textbf{1}})\right\}$, and the loss function $L^{\textbf{2}}_{IAML}$ for training $\left\{f^{\textbf{2}}_{\textbf{\textrm{CNN}}}(\bm{\theta}_{\textrm{\textbf{CNN}}}^{\textbf{2}})\right.$, $\left.f^{\textbf{2}}_{\textbf{\textrm{metric}}}(\bm{\theta}_{\textrm{\textbf{metric}}}^{\textbf{2}})\right\}$ of 3D shapes. We also developed a loss function $L_{T}$ for training the cross-modality transformation network $f_{\textbf{\textrm{trans}}}(\bm{\theta}_{\textrm{\textbf{trans}}})$ in Eq. \eqref{eq:loss_adapt}.

To learn parameters of the proposed deep cross-modality adaptation model, we optimize different networks in an alternating iterative way. Algorithm \ref{algo:CMT} summarizes the outline of the training process. Specifically, we first pre-train the CNN and metric networks of sketches and 3D shapes based on the loss $L^{m}_{IAML}$ in Eq. \eqref{eq:loss_metric_learning}, and pre-train the cross-modality transformation network by minimizing $L_{T}$ and $L_{D}$. After initialization, we then alternatively update $\left\{\bm{\theta}_{\textrm{\textbf{CNN}}}^{\textbf{1}}, \bm{\theta}_{\textrm{\textbf{metric}}}^{\textbf{1}}\right\}$, $\left\{\bm{\theta}_{\textrm{\textbf{CNN}}}^{\textbf{1}}, \bm{\theta}_{\textrm{\textbf{metric}}}^{\textbf{1}}\right\}$, $\bm{\theta}_{\textrm{\textbf{trans}}}$, and the adversarial discriminator $D$, by minimizing $L^{\textbf{1}}_{IAML}$, $L^{\textbf{2}}_{IAML}$, $L_{T}$, and $L_{D}$, respectively. Throughout the whole training process, we use the Adam stochastic gradient method \cite{kingma2014adam} as the optimizer.

\section{Experimental Results and Analysis}
\begin{figure}[!t]
\begin{center}
\centering
\subfigure[]{ 
\includegraphics[width=12cm,height=2.0cm]{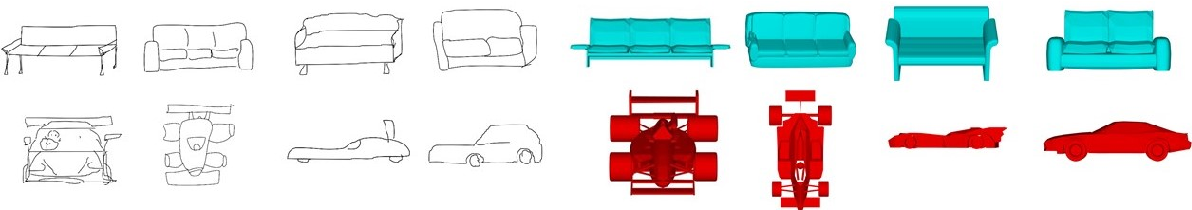}}
\hspace{0.01in}
\subfigure[]{ 
\includegraphics[width=12cm,height=2.0cm]{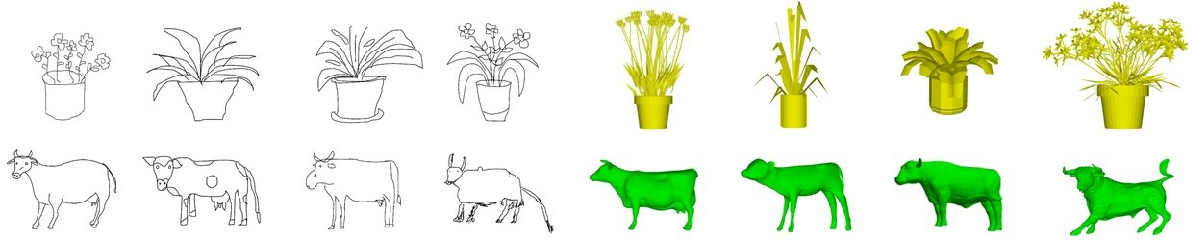}}
\end{center}
\vspace{-0.2in}
\caption{Samples from two benchmarks: (a) the \textbf{SHREC 2013} dataset, (b) the \textbf{SHREC 2014} dataset. Images in the first three columns are sketches, whilst images in the last three columns are 3D shapes. Samples in the same row belong to the same class.}\label{fig:dataset}
\vspace{-0.1in}
\end{figure}

To evaluate the performance of our method, we conduct experiments on two widely used benchmark datasets for sketch-based 3D shape retrieval: \ie, \textbf{SHREC 2013} and \textbf{SHREC 2014}.

\noindent \textbf{SHREC 2013} \cite{li2013shrec,li2014comparison} is a large-scale dataset for sketch-based 3D shape retrieval. This dataset consists of 7,200 sketches and 1,258 shapes from 90 classes, by collecting human-drawn sketches \cite{eitz2012sketch} and 3D shapes from the Princeton Shape Benchmark (PSB) \cite{shilane2004princeton} that share common categories. For each class, there are totally 80 sketches, where 50 images are used for training and 30 images for test. The numbers of 3D shapes are different for distinct classes, about 14 on average.

\noindent \textbf{SHREC 2014} \cite{li2014shrec,li2015comparison} is a sketch track benchmark larger than SHREC 2013. It totally contains 13,680 sketches and 8,987 3D shapes, grouped into 171 classes. The 3D shapes are collected from various datasets, including SHREC 2012 \cite{li2012shrec} and the Toyohashi Shape
Benchmark (TSB) \cite{tatsuma2012large}. Similar to SHREC 2013, there are 80 images for sketches, and about 53 3D shapes on average for each class. The sketches are further split into 8,550 training data and 5,130 test data, where for each class, 50 images are used for training and the rest 30 images for test.

Fig. \ref{fig:dataset} shows some samples from the two datasets. As illustrated, retrieving 3D shapes by sketches is quite challenging, due to large intra-class variations and cross-modality discrepancies between sketches and 3D shapes.

\subsection{Implementation Details}
In this subsection, we will provide implementation details about our proposed method, including network structures and parameter settings.

\noindent \textbf{Network Structures.} For CNN networks of both sketches and shapes, \ie, $f^{\textbf{1}}_{\textrm{\textbf{CNN}}}$ and $f^{\textbf{2}}_{\textrm{\textbf{CNN}}}$, we utilize the ResNet-50 network \cite{He2015}. Specifically, we use the layers of ResNet-50 before the ``\emph{pooling5}" layer (inclusive). As for metric networks of sketches and 3D shapes, \ie,  $f^{\textbf{1}}_{\textrm{\textbf{metric}}}$ and $f^{\textbf{2}}_{\textrm{\textbf{metric}}}$, both of them consist of four fully connected layers set as 2048-1024-512-256-128. We utilize the ``\emph{relu}" activation functions and batch normalization for all layers in the metric networks, except that the last layer uses the ``\emph{tanh}" activation function. As to the cross-modality transformation model $f_{\textrm{\textbf{trans}}}$, we adopt a network with four fully connected layers set as 128-64-32-64-128, where the first three layers uses the ``\emph{relu}" activation functions, and the last layer uses the ``\emph{tanh}" activation function. The discriminator $D$ is a fully connected network set as 128-64-1.

\noindent \textbf{Parameter Settings.} We set the number of the maximal iterative step $Iter_{\max}$ as 30,000. The initial learning rate is set to $1\times 10^{-4}$, and decays exponentially after 10,000 steps. To generate data batches $\bm{\mathcal{I}}^{\textbf{1}}$ and $\bm{\mathcal{I}}^{\textbf{2}}$, the number of classes $C$ per batch and the number of images $K$ per class are set as 16 and 4, respectively.

\subsection{Evaluation Metrics}
We adopt the most widely used metrics for sketch-based 3D shape retrieval as follows: nearest neighbor (\textbf{NN}), first tier (\textbf{FT}), second tier (\textbf{ST}), E-measure (\textbf{E}), discounted cumulated gain (\textbf{DCG})
and mean average precision (\textbf{mAP}) \cite{li2014comparison,dai2017deep,CVPR2017Xie}. We also report the \textbf{precision-recall curve}, a common metric for visually evaluating the retrieval performance.

\subsection{Evaluation of the Proposed Method}

In this section, we will evaluate the effect of the proposed adversarial learning with class-aware cross-modality mean discrepancy minimization (CMDM-AL), together with the semantics preserving (SeP) term.

As a baseline, we apply the importance-aware metric learning (IAML) to separately train $\{f^{\textbf{1}}_{\textbf{CNN}},f^{\textbf{1}}_{\textbf{metric}}\}$ for 2D sketches, and $\{f^{\textbf{2}}_{\textbf{CNN}},f^{\textbf{2}}_{\textbf{metric}}\}$ for 3D shapes, where the learnt feature vectors are directly used for retrieval. This baseline method, denoted by \textbf{sepIAML}, merely learns discriminative features, without considering the cross-modality issues. Based on sepIAML, we employ the cross-modality transformation network $f_{\textbf{trans}}$, which is trained by minimizing the loss $L_{AL}$. We denote this method by \textbf{DCA (CMDM-AL)}. By further adding the semantics preserving term $L_{SeP}$, \ie, training $f_{\textbf{trans}}$ by $L_{T}=L_{AL}+L_{SeP}$, we can obtain the complete model of our proposed method denoted by \textbf{DCA (CMDM-AL+SeP)}. By comparing the performance of sepIAML, DCT (CMDM-AL) and DCT (CMDM-AL+SeP), we can evaluate the effects of the proposed adversarial learning method and semantics preserving term.

The results are summarized in Tables \ref{table:res_SHREC13} and \ref{table:res_SHREC14}. As can be seen, the baseline method sepSMML yields a rather poor performance, due to its weakness in dealing with cross-modality discrepancies. By introducing the adversarial learning method, DCA (CMDM-AL) significantly boosts the performance of the baseline, implying that the adversarial learning can largely enhance the correlation between data distributions of different modalities. Moreover, we can see a consistent improvements of DCA (CMDM-AL+SeP) on two benchmarks, compared to DCA (CMDM-AL). This indicates that the semantics preserving term can help learn more discriminative cross-modality transformation network.

\subsection{Comparison with the State-of-the-art Methods}

\noindent \textbf{Retrieval Performance on SHREC 2013.} Here we report experimental results of the proposed method on SHREC 2013, by comparing with the state-of-the-art methods, including the cross domain manifold ranking method (\textbf{CDMR}) \cite{furuya2013ranking}, sketch-based retrieval method with view clustering (\textbf{SBR-VC}) \cite{li2013shrec}, spatial proximity method (\textbf{SP}) \cite{sousa2010sketch}, Fourier descriptors on 3D model silhouettes (\textbf{FDC}) \cite{li2013shrec}, edge-based Fourier spectra descriptor (\textbf{EFSD}) \cite{li2013shrec}, Siamese network (\textbf{Siamese}) \cite{wang2015sketch}, chordal axis transform with dynamic time warping (\textbf{CAT-DTW}), deep correlated metric learning (\textbf{DCML}) \cite{dai2017deep}, and the learned Wasserstein barycentric representation method (\textbf{LWBR}) \cite{CVPR2017Xie}.

\setlength{\tabcolsep}{4pt}
\begin{table}[!t]
\begin{center}
\caption{Performance on \textbf{SHREC 2013}, compared with the state-of-the-art methods. (The symbol ``NA" in this table means that the data is not available.)}
\label{table:res_SHREC13}
\begin{tabular}{lllllll}
\hline\noalign{\smallskip}
\textbf{Methods} & \textbf{NN} & \textbf{FT} & \textbf{ST} & \textbf{E} & \textbf{DCG} & \textbf{mAP}\\
\noalign{\smallskip}
\hline
\noalign{\smallskip}
CDMR \cite{furuya2013ranking} & 0.279 & 0.203 & 0.296 & 0.166 & 0.458 & 0.250\\
SBR-VC \cite{li2013shrec}     & 0.164 & 0.097 & 0.149 & 0.085 & 0.348 & 0.114\\
SP \cite{sousa2010sketch}     & 0.017 & 0.016 & 0.031 & 0.018 & 0.240 & 0.026\\
FDC \cite{li2013shrec}        & 0.110 & 0.069 & 0.107 & 0.061 & 0.307 & 0.086\\
Siamese \cite{wang2015sketch} & 0.405 & 0.403 & 0.548 & 0.287 & 0.607 & 0.469\\
CAT-DTW \cite{yasseen2017view}& 0.235 & 0.135 & 0.198 & 0.109 & 0.392 & 0.141\\
KECNN \cite{tabia2017learning}& 0.320 & 0.319 & 0.397 & 0.236 & 0.489 & NA\\
DCML \cite{dai2017deep}       & 0.650 & 0.634 & 0.719 & 0.348 & 0.766 & 0.674\\
LWBR \cite{CVPR2017Xie}       & 0.712 & 0.725 & 0.785 & 0.369 & 0.814 & 0.752\\
\hline
Baseline (sepIAML)            & 0.011 & 0.015 & 0.028 & 0.016 & 0.234 & 0.037\\
\textbf{DCA (CMDM-AL)}        & 0.762 & 0.776 & 0.812 & 0.370 & 0.842 & 0.795\\
\textbf{DCA (CMDM-AL+SeP)}    & \textbf{0.783} & \textbf{0.796} & \textbf{0.829} & \textbf{0.376} & \textbf{0.856} & \textbf{0.813}\\
\hline
\end{tabular}
\end{center}
\end{table}
\setlength{\tabcolsep}{1.4pt}

\begin{figure}[!t]
\centering
\includegraphics[width=16cm]{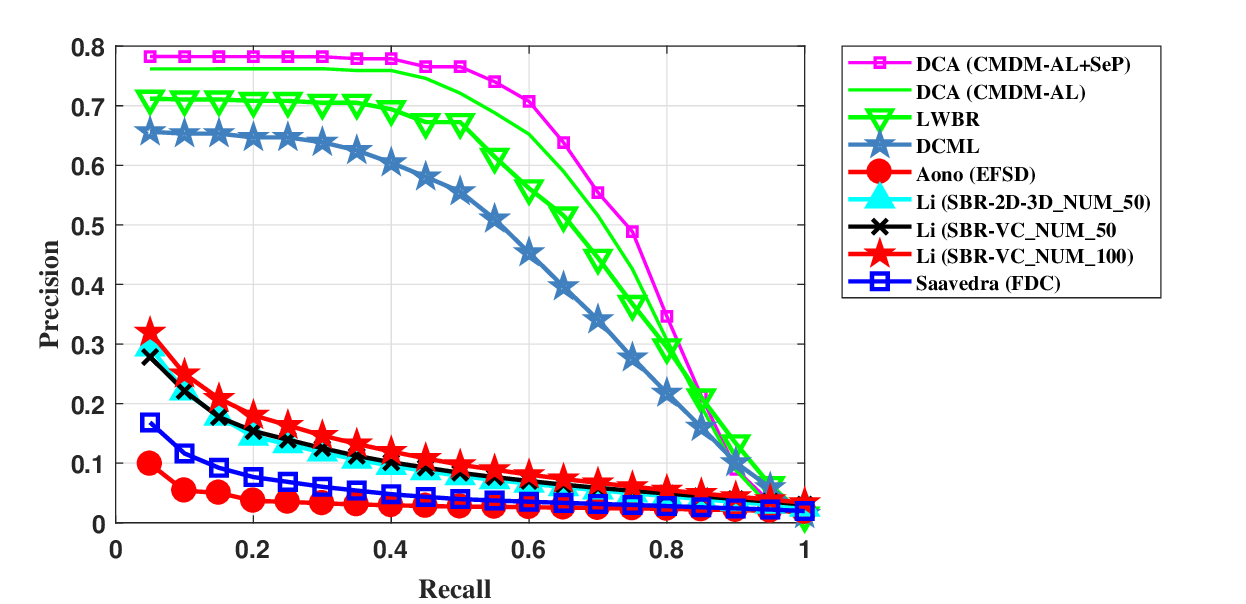}
\caption{The precision-recall curves of various methods on \textbf{SHREC 2013}.}
\label{fig:pre_rec_curve_shrec13}\vspace{-0.1in}
\end{figure}

Fig. \ref{fig:pre_rec_curve_shrec13} demonstrates the precision-recall curves of the proposed method and compared approaches. As illustrated, the precision rate of our method is significantly higher than those of compared models, when the recall rate is smaller than 0.8. Considering that the top retrieved results are preferable, our method therefore performs significantly better than the state-of-the-art approaches.

We also report NN, FT, ST, E, DCG and mAP of various methods, including CDMR, SBR-VC, SP, FDC, Siamese, DCML, LWBR and the proposed method. As summarized in Table \ref{table:res_SHREC13}, our approach yields the best retrieval performance w.r.t. all evaluation metrics. Among all compared approaches, Siamese, DCML and LWBR are deep metric learning based models. They directly map data from different modalities into a common embedding subspace, where both the single-modality and cross-modality intra-class Euclidean distances are decreased, and the inter-class distances are simultaneously enlarged. However, they equally treat each training data, and fail to explore varying importance of distinct samples. Besides, they only reduce the local cross-modality divergences between data pairs or triplets, without considering the correlation between data distributions in a holistic way. In contrast, our method learns features by mining the batch-wise hardest positive and hardest negative samples. Through automatically selecting the most important training samples, we can learn discriminative features more efficiently. Moreover, we explicitly introduce a cross-modality transformation network, in order to transfer the feature from the sketch modality to the feature space of 3D shapes. By leveraging the semantics preserving adversarial learning, we simultaneously reduce holistic divergences between data distributions from two modalities, and enhance the semantic correlations. As a consequence, our method achieves better retrieval performance. For instance, the mAP of our method reaches 0.813, which is $34.4\%$, $13.9\%$ and $6.1\%$ higher than Siamese, DCML and LWBR, respectively.

\noindent \textbf{Retrieval Performance on SHREC 2014.} On this dataset, we compared our proposed model to the following state-of-the-art methods: the BoF with Gabor local line based feature (\textbf{BF-fGALIF})\cite{eitz2012sketch}, \textbf{CDMR} \cite{furuya2013ranking}, \textbf{SBR-VC} \cite{li2013shrec}, depth-buffered vector of locally
aggregated tensors (\textbf{DB-VLAT}) \cite{tatsuma2012large}, \textbf{SCMR-OPHOG} \cite{li2014shrec}, BOF junction-based extended shape context (\textbf{BOFJESC}) \cite{li2014shrec}, \textbf{Siamese} \cite{wang2015sketch}, \textbf{DCML} \cite{dai2017deep}, and \textbf{LWBR} \cite{CVPR2017Xie} .

\setlength{\tabcolsep}{4pt}
\begin{table}[!t]
\begin{center}
\caption{Performance on \textbf{SHREC 2014}, compared with the state-of-the-art methods}
\label{table:res_SHREC14}
\begin{tabular}{lllllll}
\hline\noalign{\smallskip}
\textbf{Methods} & \textbf{NN} & \textbf{FT} & \textbf{ST} & \textbf{E} & \textbf{DCG} & \textbf{mAP}\\
\noalign{\smallskip}
\hline
\noalign{\smallskip}
CDMR \cite{furuya2013ranking}                & 0.109 & 0.057 & 0.089 & 0.041 & 0.328 & 0.054\\
SBR-VC \cite{li2013shrec}                    & 0.095 & 0.050 & 0.081 & 0.037 & 0.319 & 0.050\\
DB-VLAT \cite{tatsuma2012large}              & 0.160 & 0.115 & 0.170 & 0.079 & 0.376 & 0.131\\
CAT-DTW \cite{yasseen2017view}               & 0.137 & 0.068 & 0.102 & 0.050 & 0.338 & 0.060\\
Siamese \cite{wang2015sketch}                & 0.239 & 0.212 & 0.316 & 0.140 & 0.496 & 0.228\\
DCML \cite{dai2017deep}                      & 0.272 & 0.275 & 0.345 & 0.171 & 0.498 & 0.286\\
LWBR \cite{CVPR2017Xie}                      & 0.403 & 0.378 & 0.455 & 0.236 & 0.581 & 0.401\\
\hline
Baseline (sepIAML)                           & 0.016 & 0.016 & 0.023 & 0.005 & 0.263 & 0.028\\
\textbf{DCA (CMDM-AL)}                       & 0.745 & 0.766 & 0.808 & 0.392 & 0.845 & 0.782\\
\textbf{DCA (CMDM-AL+SeP)}                   & \textbf{0.770} & \textbf{0.789} & \textbf{0.823} & \textbf{0.398} & \textbf{0.859} & \textbf{0.803}\\
\hline
\end{tabular}
\end{center}
\end{table}

\begin{figure}[!t]
\centering
\includegraphics[width=16cm]{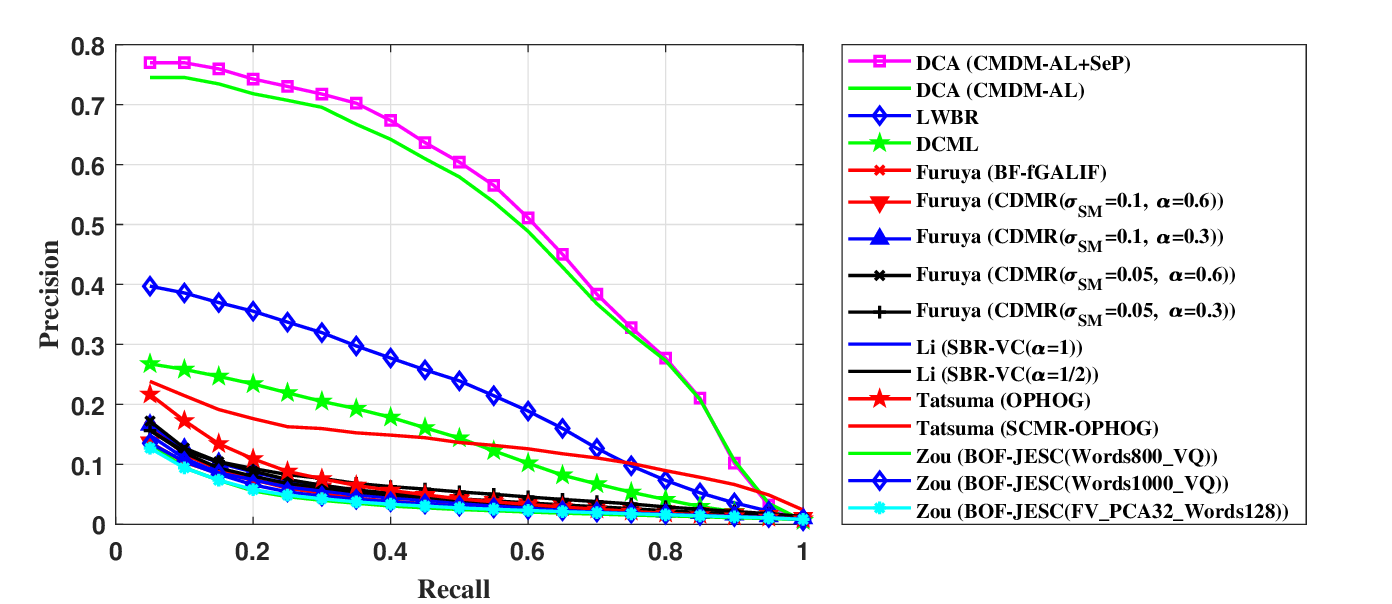}
\caption{The precision-recall curves of various methods on \textbf{SHREC 2014}.}
\label{fig:pre_rec_curve_shrec14}
\vspace{-0.2in}
\end{figure}

Fig. \ref{fig:pre_rec_curve_shrec14} provides precision-recall curves for BF-fGALIF, CDMR, SBR-VC, SCMR-OPHOG, OPHOG, DCML, LWBR and the proposed model. As shown, the precision rate of our proposed method is remarkably higher than compared approaches, when the recall rate is less than 0.8.

Besides the precision-recall curves, we additionally report NN, FT, ST, E, DCG and mAP of CDMR, SBR-VC, DB-VLAT, Siamese, DCML, LWBR in Table \ref{table:res_SHREC14}. As can be seen, the performance of existing deep metric learning based methods including Siamese, DCML and LWBR drops sharply on SHREC 2014. For example, the mAP of LWBR on SHREC 2014 is 0.401, around $35\%$ lower than the mAP that it has achieved on SHREC 2013. The reason might lie in that SHREC 2014 has much more class categories (90 classes on SHREC 2013 versus 171 classes on SHREC 2014) and larger scale 3D shapes (1,258 3D shapes on SHREC 2013 versus 8,987 3D shapes on SHREC 2014) with more severe intra-class and cross-modality variations, making SHREC 2014 more challenging than SHREC 2013. As a comparison, the mAP of our proposed model merely drops about $1\%$, and reaches 0.803 on SHREC 2014. This result is 40.2\%, 51.7\% and 57.5\% higher than that of LWBR, DCML and Siamese, indicating that our method are much more scalable than existing deep models. 

\vspace{-0.2in}

\section{Conclusions}

In this paper, we proposed a novel cross-modality adaptation model for sketch-based 3D shape retrieval. We firstly learnt modality-specific discriminative features for 2D sketches and 3D shapes, by employing the importance-aware metric learning through mining the batch-wise hardest samples. To remove the cross-modality discrepancy, we proposed a transformation network, aiming to transfer the features of sketches into the feature space of 3D shapes. We developed an adversarial learning based method for training the network, by enhancing correlations between holistic data distributions and preserving local semantic structures across modalities. As a consequence, we obtained discriminative transformed features of sketches that were also highly correlated with data of 3D shapes. Extensive experimental results on two benchmark datasets demonstrated the superiority of the propose method, compared to the state-of-the-art approaches.

{\small
\bibliographystyle{ieee}
\bibliography{egbib}
}
\end{document}